\documentclass[tablecaption=bottom,wcp]{jmlr} 
\usepackage{booktabs}
\usepackage{xcolor}
\usepackage{svg}
\usepackage{multirow}

\usepackage[load-configurations=version-1]{siunitx} 

\theorembodyfont{\upshape}
\theoremheaderfont{\scshape}
\theorempostheader{:}
\theoremsep{\newline}

\jmlrproceedings{AABI 2020}{3rd Symposium on Advances in Approximate Bayesian Inference, 2020}

\title[Variational Determinant Estimation]{Variational Determinant Estimation \\ with Spherical Normalizing Flows}

\author{\Name{Simon Passenheim} \Email{simon.passenheim@gmail.com} \\
 \Name{Emiel Hoogeboom} \Email{e.hoogeboom@uva.nl}\\
 \addr University of Amsterdam}

\begin{document}

\maketitle

\begin{abstract}
This paper introduces the \textit{Variational Determinant Estimator} (VDE), a variational extension of the recently proposed determinant estimator discovered by \citet{sohl2020two}. Our estimator significantly reduces the variance even for low sample sizes by combining (importance-weighted) variational inference and a family of normalizing flows which allow density estimation on hyperspheres. In the ideal case of a tight variational bound, the VDE becomes a zero variance estimator, and a single sample is sufficient for an exact (log) determinant estimate.
\end{abstract}



\section{Introduction}
\label{sec:intro}
\begin{figure}[b]
\centering
\includegraphics[width=1\textwidth]{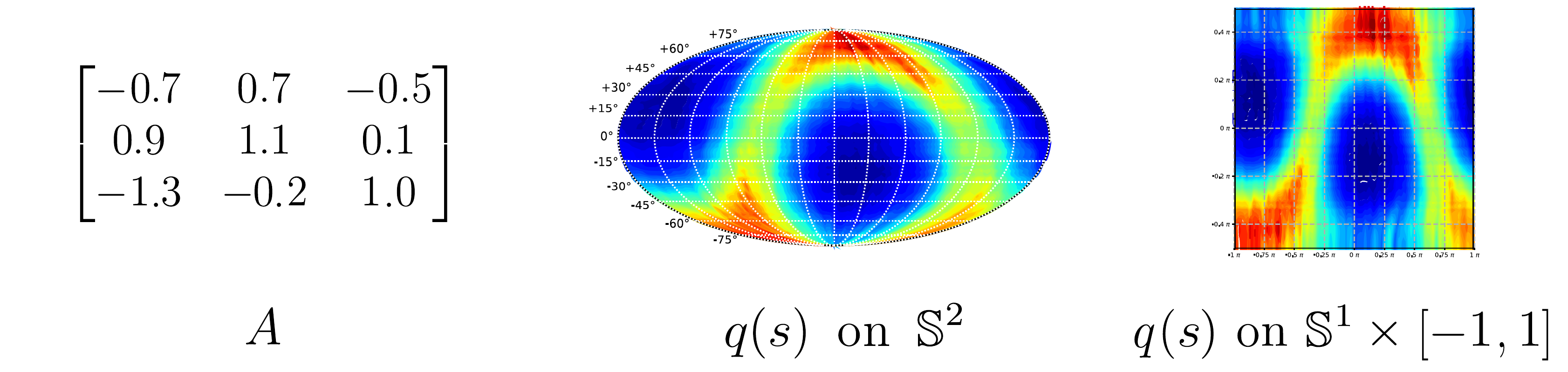}
\vspace{-.5cm}
\caption{Overview of variational determinant estimation. A spherical proposal distribution $q(s)$ is trained to estimate the absolute determinant of a matrix $A$.}
\label{fig:overview}
\end{figure}

The computation of the (log) absolute determinant of matrices is a problem that is encountered in machine learning in areas such as normalizing flows \citep{rezende2015variational, dinh2016density}.
\cite{sohl2020two} connects the inverse absolute determinant of $A$ with an expectation over matrix-vector products:  
\begin{align}
\label{eq:det_expactation}
    \lvert A \rvert^{-1} = \mathbb{E}_{s \sim \mathcal{U}(\mathbb{S}^{n-1})} \left[\lVert A s \rVert^{-n} \right], 
 \end{align}
where samples are drawn from a uniform distribution on the $n-1$ dimensional hypersphere. For better readibility we will write shorthand $\mathcal{U}(s)$ for the uniform spherical distribution in the following. A common unbiased estimator for Equation \ref{eq:det_expactation} is then given via Monte Carlo (MC) integration:
\begin{align}
\label{eq:det_expactation_estimate}
 \mathbb{E}_{s \sim \mathcal{U}(s)} \left[ \lVert A s \rVert^{-n} \right] \approx \frac{1}{N} \sum_{i=1}^N \lVert A s_i \rVert^{-n} \quad \text{with} \quad s_i \sim \mathcal{U}(s).
\end{align}
\cite{sohl2020two} has empirically shown that the na\"{i}ve MC approach of Equation \ref{eq:det_expactation_estimate} can have problematic high variance, meaning that we need around $10^6$ samples to correctly estimate even a $10 \times 10$ matrix.

This paper extends the MC determinant estimator using Spherical Normalizing Flows to introduce the Variational Determinant Estimator. This new estimator achieves lower variance and, as a result, requires fewer samples for accurate estimates. For an example see Figure \ref{fig:overview}.

\subsection{Spherical Normalizing Flows}
Normalizing flows \citep{rezende2015variational, dinh2016density, kingma2018glow} are generative models which transform a base distribution $\pi(z)$ on a space $Z$ into a more complex distribution $q(s)$ on another space $S$ via a diffeomorphism $f \colon Z \to S$. The relationship between those is given by the change of variables formula:
\begin{align}
\label{eq:change_of_variables}
    \log q(s) = \log \pi(z) - \log \lvert \det J_{f}(z) \rvert,
\end{align}
where $J_{f}$ is the Jacobian. Most work has been done when $Z$ and $S$ are Euclidean spaces with flat geometry. Flows for hyperspherical geometries were introduced in \cite{rezende2020normalizing}, that is $f \colon \mathbb{S}^n \to \mathbb{S}^n$ is a diffeomophism. For these flows we usually choose a uniform base distribution $\pi(z) = \mathcal{U}(z)$ since the underlying spaces are compact. As a consequence, $q(s)$ is a distribution on the hypersphere parametrized by complicated invertible functions that we can straightforwardly sample from and compute the likelihood.

\section{The Variational Determinant Estimator}
\label{sec:method}
To estimate the determinant and log determinant more efficiently with less variance, we introduce the Variational Determinant Estimator:
\begin{equation*}
    \lvert A \rvert^{-1} = \mathbb{E}_{s \sim \mathcal{U}(s) }\left[\lVert A s \rVert^{-n} \right] = \mathbb{E}_{s \sim q(s)} \left[ \frac{\mathcal{U}(s)}{q(s)} \lVert A s \rVert^{-n}  \right],
 \end{equation*}
 which by \cite{owen2013monte} has the least variance when $q(s) \propto \mathcal{U}(s) \lVert As\rVert^{-n}$. An example to achieve this proportionality is to minimize the divergence
 \begin{equation*}
\mathrm{KL}(q(s) \: ; \: \mathcal{U}(s) \lVert As\rVert^{-n} / Z),
\end{equation*} 
where $\mathcal{U}(s) \lVert As\rVert^{-n}$ is treated as an (unnormalized) probability distribution and $Z$ is an unknown normalization constant which does not influence the gradient. To avoid clutter with unnecessary constants, we drop $Z$ in the following and note that the resulting KL divergence is an abuse of notation because the well-known properties such as non-negativity do not necessarily hold anymore for $\mathrm{KL}(q(s) \: ; \: \mathcal{U}(s) \lVert As\rVert^{-n} )$. Using this divergence has the additionally desired effect that:
\begin{equation*}
\resizebox{1\hsize}{!}{
$\log \lvert A \rvert^{-1} = \log \mathbb{E}_{s \sim q(s)} \left[ \frac{\mathcal{U}(s)}{q(s)} \lVert A s \rVert^{-n} \right] \geq \mathbb{E}_{s \sim q(s)} \left[ \log \frac{\mathcal{U}(s)}{q(s)} \lVert A s \rVert^{-n} \right] = - \mathrm{KL}(q(s) \: ; \: \mathcal{U}(s) \lVert As\rVert^{-n})$,
}
\end{equation*}
and thus the negative $\mathrm{KL}$ gives a lower bound on the $\log$ absolute determinant of $A^{-1}$ due to Jensen's inequality. Consequently, this gives a direct method to estimate the $\log$ absolute determinant of $A$ using the upper bound $\mathrm{KL}(q(s) \: ; \: \mathcal{U}(s) \lVert As\rVert^{-n})$ which is tight when $q(s) \propto \mathcal{U}(s) \lVert  As\rVert^{-n}$. In this ideal case, the VDE becomes a zero variance estimator, see \citet{golinski2019amortized} Section 2.1.

The proposal distribution $q$ is modeled by a flow introduced in Section \ref{sec:intro}. 
If we substitute $q$ in $\mathrm{KL}$ using the change of variables formula in Equation \ref{eq:change_of_variables} with a normalizing flow $f$, and we choose a uniform base distribution $\pi(s) = \mathcal{U}(s)$, the objective simplifies:
\begin{align}
\label{eq:simple_objective}
\begin{split}
    \mathrm{KL}(q(s) \: ; \: \mathcal{U}(s) \lVert As\rVert ^{-n}) &= \mathbb{E}_{s \sim q(s)} \left[\log q(s) - \log  \lVert As\rVert^{-n} - \log \mathcal{U}(s)\right]\\
    &= \mathbb{E}_{s_0 \sim \mathcal{U}(s)} \left[ - \log \vert \det J_f(s_0) \rvert + n \log \lVert A f(s_0)\rVert \right].
\end{split}
\end{align}
Equation \ref{eq:simple_objective} is optimized via naive Monte Carlo integration. The result of such a learned $q$ in the case of a $3 \times 3$ matrix $A$ is illustrated in Figure \ref{fig:overview} and the optimal proposal distribution is visualized in Figure \ref{fig:opt_proposal} of the Appendix \ref{sec:app_cover_matrix}.

\section{Related Work}
\label{sec:rel_work}
Importance sampling has a long history as a study object. \citet{hesterberg1988advances} introduced extensions for the importance weights such as regression or non-linear exponential estimates to allow the method to be effectively applied in a wider range of settings such as multivariate outputs. \citet{kingmavae,rezende2014stochastic} have introduced deep learning-based variational inference and \citet{burda2015importance} have shown tighter bounds with importance-weighted variational inference. Although these works were originally aimed at estimating log probabilities, they can be more generally be applied to marginalize a probabilistic latent variable. \citet{muller2019neural} utilize flows to learn a proposal distribution for importance sampling in a Euclidean space. \citet{golinski2019amortized} introduce amortized Monte Carlo integration, which combines different proposal distributions for better estimates.

Normalizing Flows \citep{tabak2013family,rezende2015variational} are an attractive generative model to learn distributions because they admit exact likelihood evaluation, and they are fast to sample from as opposed to autoregressive models. There have been many advances for flows on Euclidean manifolds \citep{dinh2016density,kingma2016improved,chen2019residual,perugachi2020idensenets}. Recently, \citet{gemici2016normalizing,rezende2020normalizing} have introduced normalizing flows for hyperspheres. As a result, it is now also possible to learn expressive distributions on hyperspheres with exact likelihood estimates and efficient sampling.

\section{Results}
\label{sec:results}
In this section, we demonstrate the performance of our method in determinant estimation. We consider two cases: estimating the determinant of randomly sampled $10 \times 10$ dense matrices and estimating the determinant of a convolutional layer.

\begin{table}
\centering
\begin{tabular}{lccccc} 
\toprule
Nr. of samples & $10^2$ & $10^3$ & $10^4$ & $10^5$ \\
\midrule
VDE det. (ours) & \textbf{3.4} {\small$\pm$ 2.1} \%  & \textbf{1.7} {\small $\pm$ 0.6} \% & \textbf{1.6} {\small $\pm$ 1.3} \% & \textbf{0.3} {\small $\pm$ 0.3} \% \\
MC det. & 533 {\small $\pm$ 660 } \% & 348 \small{$\pm$ 262} \% & 104 \small{$\pm$ 30} \% &  59 \small{$\pm$  43} \%  \\ 
\bottomrule
\end{tabular}
\caption{Comparison of determinant estimates. Results are in mean absolute relative difference for $5$ unit Gaussian sampled $10 \times 10$ matrices. Deviations are given in one standard deviation.}
\label{tab:averaged_rel_diff}
\end{table}

\subsection{Dense Matrix}
\label{sec:dense_matrix}
The determinant is estimated for five $10 \times 10$ matrices where the entries are sampled from unit Gaussians, see the Appendix \ref{sec:app_dense_matrix} for the specific matrices. The spherical flow utilizes a Moebius transformation for the circle part with $N_C$ = 12 number of centers and Neural Spline flows \citep{durkan2019neural} with $N_B = 16$ number of bins for the interval part, see \cite{rezende2020normalizing} for details of the architecture and the parameters. Furthermore, we stacked $N_F=8$ flows on top, used coupling layers, and trained the models for $10k$ iterations with a batch size of $1024$. Flows based on autoregressive masking are also possible.

The results can be seen in Table \ref{tab:averaged_rel_diff}, where we present the mean of the relative absolute differences of the estimated determinant in comparison to the true absolute determinant. The variational determinant estimator achieves even for a low sample size of $10^2$ low relative differences, whereas in contrast, the na\"{i}ve Monte Carlo estimate still has high variance throughout all sample sizes.

\begin{figure}
\centering
\subfigure[Absolute determinant estimates.]{\includegraphics[width=200pt]{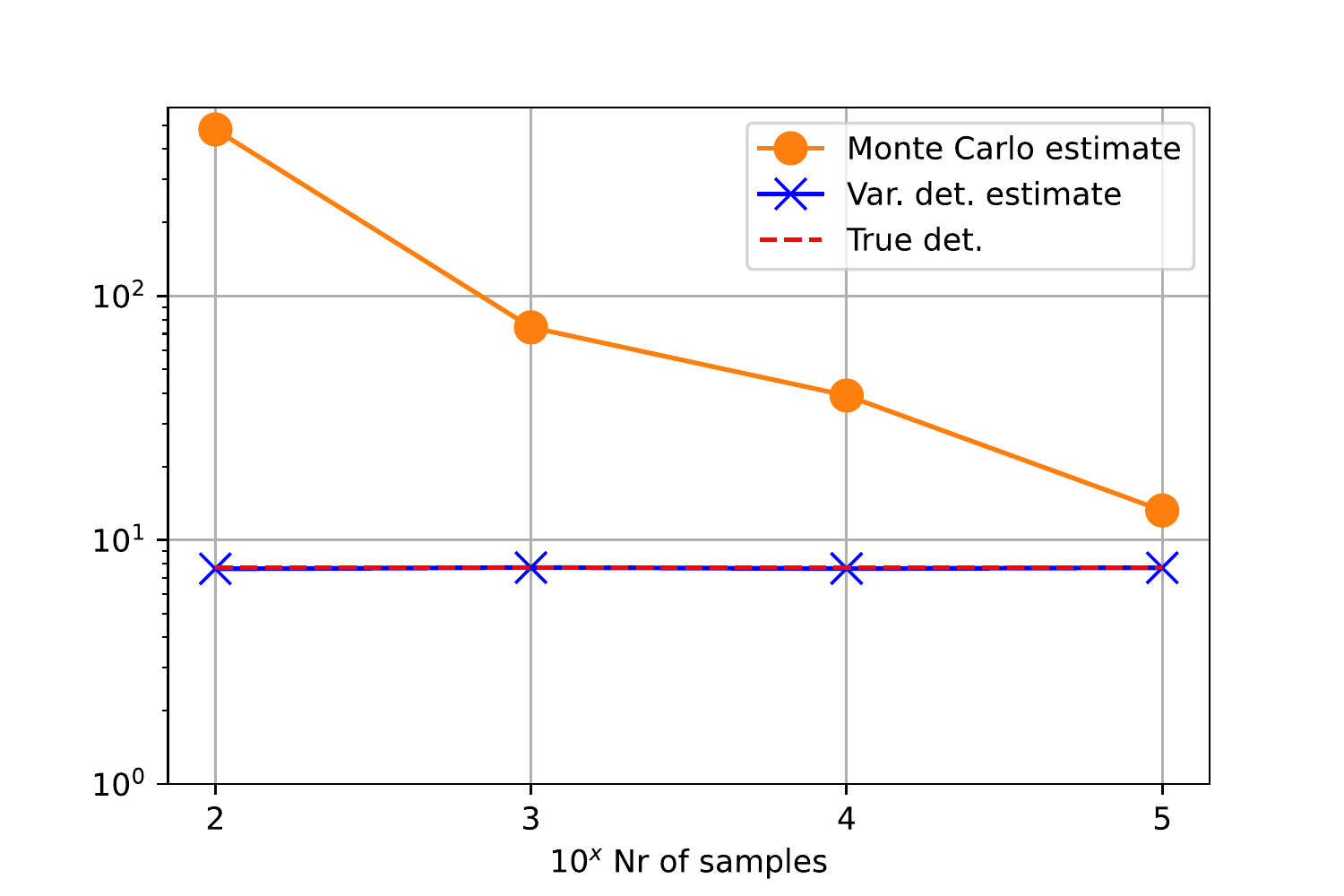}\label{fig:det_estimates}}
\hspace{0.5cm}
\subfigure[Relative variation of determinant estimates]{\includegraphics[width=200pt]{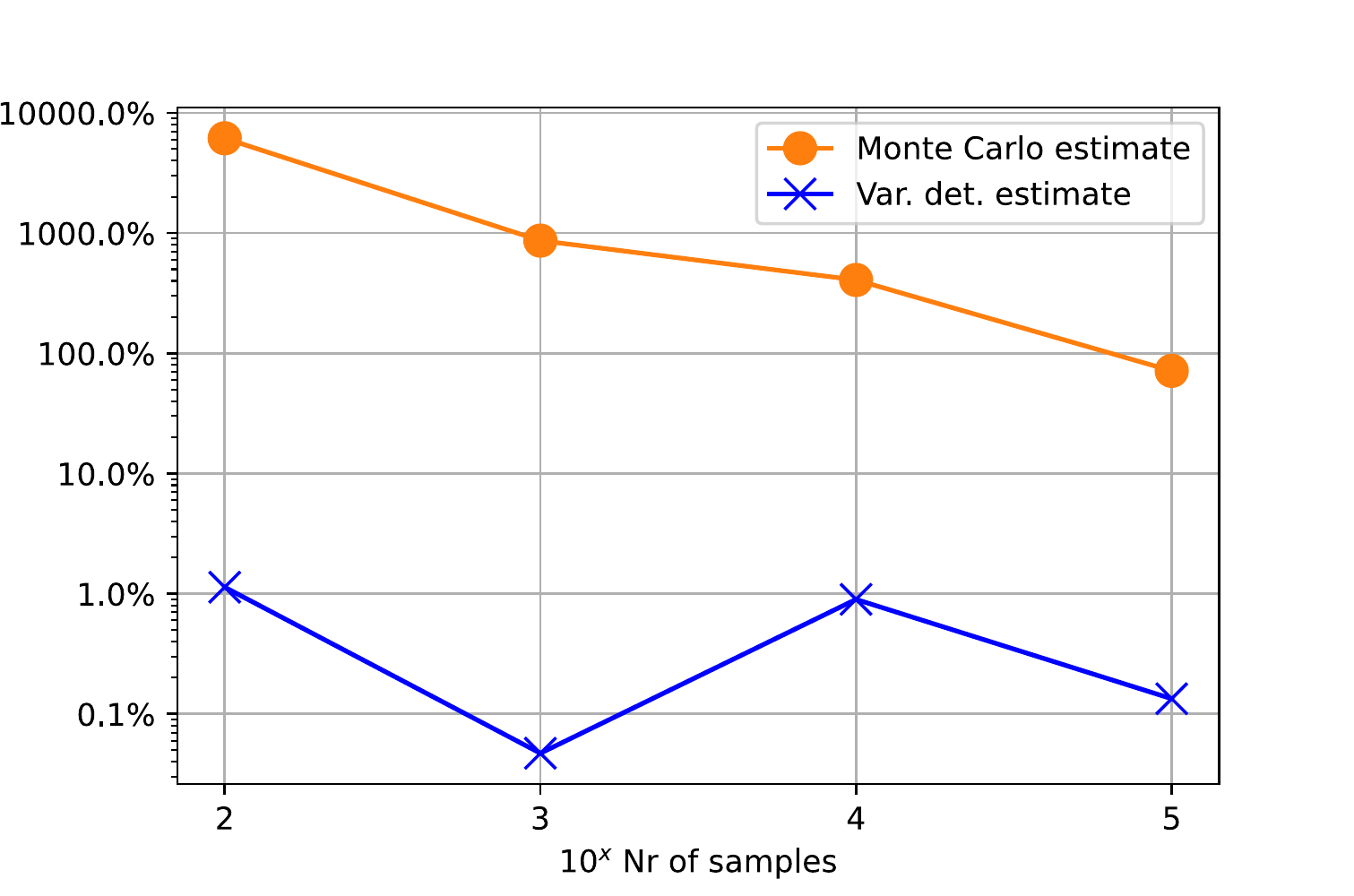}\label{fig:relative_diff}}
\caption{Determinant estimates of a structured $16 \times 16$ matrix. Left: absolute determinant estimated values. Right: the absolute relative difference to the true determinant value. Both plots are on log-scale.}
\label{fig:conv_exp}
\vspace{-.3cm}
\end{figure}

\begin{table}[ht]
\centering
\begin{tabular}{lrrrrc} 
\toprule
Nr. of samples &  $10^2$ & $10^3$ & $10^4$ & $10^5$  & true (log) determinant \\  
\midrule 
VDE det. (ours) & 7.62 & 7.71 & 7.64 & 7.70 &  \multirow{2}{*}{7.71} \\
MC det.  & 481.09 &  74.37 &  39.08 &  13.21 & \\
\midrule 
VDE log det. (ours)  & 2.03 & 2.04 & 2.03 & 2.04 &  \multirow{2}{*}{2.04} \\
MC log det. & 6.18 &  4.31 &  3.67 & 2.58 & \\
\midrule
VDE Rel. diff of det. (ours)  & \textbf{1.1} \% & \textbf{0.05} \% & \textbf{0.9} \% & \textbf{0.1} \% & \multirow{2}{*}{0 \%} \\
MC Rel. diff of det. & 6144 \% &  865 \% &  407 \% &  71 \% &  \\ 
\bottomrule
\end{tabular}
\caption{Comparison of variational and Monte Carlo determinant estimates. First four rows show results in absolute numbers and log-space. The last two rows illustrate the relative difference of the estimates to the true determinant.}
\label{tab:conv_exp}
\end{table}

\subsection{Convolutional Layer}
\label{sec:conv_layer} 
In this experiment, the (log) determinant of a convolutional layer is estimated. The reason for this experiment is that these types of sparse linear transformations often occur in deep learning, and they typically have cheap matrix-vector products. We manually reconstructed the $16 \times 16$ equivalent matrix $W$ of a convolution of a $3 \times 3 $ filter with an $ 4 \times 4$ image, see Appendix \ref{sec:app_conv_layer} for the filter. The determinant of $W$ equals the determinant of the convolution operation. We chose the same architecture as in our previous experiment but ran the experiment this time for $40k$ iterations with the same batch size. Note that the parametrization of our spherical flow is fully connected, and better optimization behavior is expected when the parametrization would be convolutional.

Figure \ref{fig:conv_exp} and Table \ref{tab:conv_exp} show the results of the experiment in terms of absolute and log abs. determinant estimates. We observe the same behavior as in the previous experiment: The variational determinant estimator achieves low errors with already low sample sizes, whereas the MC determinant estimator is not able to capture the determinant correctly with even high sample sizes.

\section{Conclusion}
\label{sec:conclusion}
In this paper, we introduced the Variational Determinant Estimator, which achieves with low sample sizes high accuracy in estimating a determinant of a linear operator. Interestingly, the estimator in its original variant and the VDE allows estimation if only matrix-vector products are available. 

In our experiments, we considered an offline setting where a large number of samples are required first to optimize the model. However, in future work, VDE could also be applied in an online, moving target settings. In this case, small updates to the matrix $A$ would only require small updates to the density model $q(s)$. A perpendicular direction for VDE would be to estimate the Jacobian determinant of a function $f$. The Jacobian would depend on the input $x$, and the density model $q(s|x)$ can be amortized. Additionally, not the entire Jacobian but only Jacobian-vector products would be required to estimate the determinant.

Code to reproduce the results and to enable further research concerning the Variational Determinant Estimator and spherical normalizing flows will be made public in the git repository\footnote{https://github.com/P4ppenheimer} of one of the authors at a later time point.

\newpage
\bibliography{main}

\begin{thebibliography}{18}
\providecommand{\natexlab}[1]{#1}
\providecommand{\url}[1]{\texttt{#1}}
\expandafter\ifx\csname urlstyle\endcsname\relax
  \providecommand{\doi}[1]{doi: #1}\else
  \providecommand{\doi}{doi: \begingroup \urlstyle{rm}\Url}\fi

\bibitem[Burda et~al.(2015)Burda, Grosse, and
  Salakhutdinov]{burda2015importance}
Yuri Burda, Roger Grosse, and Ruslan Salakhutdinov.
\newblock Importance weighted autoencoders.
\newblock \emph{arXiv preprint arXiv:1509.00519}, 2015.

\bibitem[Chen et~al.(2019)Chen, Behrmann, Duvenaud, and
  Jacobsen]{chen2019residual}
Tian~Qi Chen, Jens Behrmann, David Duvenaud, and J{\"{o}}rn{-}Henrik Jacobsen.
\newblock Residual flows for invertible generative modeling.
\newblock In \emph{Neural Information Processing Systems, NeurIPS}, pages
  9913--9923, 2019.

\bibitem[Dinh et~al.(2016)Dinh, Sohl-Dickstein, and Bengio]{dinh2016density}
Laurent Dinh, Jascha Sohl-Dickstein, and Samy Bengio.
\newblock Density estimation using real nvp.
\newblock \emph{arXiv preprint arXiv:1605.08803}, 2016.

\bibitem[Durkan et~al.(2019)Durkan, Bekasov, Murray, and
  Papamakarios]{durkan2019neural}
Conor Durkan, Artur Bekasov, Iain Murray, and George Papamakarios.
\newblock Neural spline flows.
\newblock In \emph{Advances in Neural Information Processing Systems}, pages
  7511--7522, 2019.

\bibitem[Gemici et~al.(2016)Gemici, Rezende, and
  Mohamed]{gemici2016normalizing}
Mevlana~C. Gemici, Danilo~Jimenez Rezende, and Shakir Mohamed.
\newblock Normalizing flows on riemannian manifolds.
\newblock \emph{CoRR}, abs/1611.02304, 2016.

\bibitem[Goli{\'n}ski et~al.(2019)Goli{\'n}ski, Wood, and
  Rainforth]{golinski2019amortized}
Adam Goli{\'n}ski, Frank Wood, and Tom Rainforth.
\newblock Amortized monte carlo integration.
\newblock \emph{arXiv preprint arXiv:1907.08082}, 2019.

\bibitem[Hesterberg(1988)]{hesterberg1988advances}
Timothy~Classen Hesterberg.
\newblock \emph{Advances in importance sampling}.
\newblock PhD thesis, Citeseer, 1988.

\bibitem[Kingma and Welling(2014)]{kingmavae}
Diederik~P. Kingma and Max Welling.
\newblock Auto-encoding variational bayes.
\newblock In \emph{2nd International Conference on Learning Representations,
  {ICLR}}, 2014.

\bibitem[Kingma et~al.(2016)Kingma, Salimans, Jozefowicz, Chen, Sutskever, and
  Welling]{kingma2016improved}
Diederik~P Kingma, Tim Salimans, Rafal Jozefowicz, Xi~Chen, Ilya Sutskever, and
  Max Welling.
\newblock {Improved variational inference with inverse autoregressive flow}.
\newblock In \emph{Advances in Neural Information Processing Systems}, pages
  4743--4751, 2016.

\bibitem[Kingma and Dhariwal(2018)]{kingma2018glow}
Durk~P Kingma and Prafulla Dhariwal.
\newblock Glow: Generative flow with invertible 1x1 convolutions.
\newblock In \emph{Advances in neural information processing systems}, pages
  10215--10224, 2018.

\bibitem[M{\"u}ller et~al.(2019)M{\"u}ller, McWilliams, Rousselle, Gross, and
  Nov{\'a}k]{muller2019neural}
Thomas M{\"u}ller, Brian McWilliams, Fabrice Rousselle, Markus Gross, and Jan
  Nov{\'a}k.
\newblock Neural importance sampling.
\newblock \emph{ACM Transactions on Graphics (TOG)}, 38\penalty0 (5):\penalty0
  1--19, 2019.

\bibitem[Owen(2013)]{owen2013monte}
Art~B Owen.
\newblock Monte carlo theory.
\newblock \emph{Methods and Examples}, 665, 2013.

\bibitem[Perugachi{-}Diaz et~al.(2020)Perugachi{-}Diaz, Tomczak, and
  Bhulai]{perugachi2020idensenets}
Yura Perugachi{-}Diaz, Jakub~M. Tomczak, and Sandjai Bhulai.
\newblock i-{D}ensenets.
\newblock \emph{CoRR}, abs/2010.02125, 2020.

\bibitem[Rezende and Mohamed(2015)]{rezende2015variational}
Danilo~Jimenez Rezende and Shakir Mohamed.
\newblock Variational inference with normalizing flows.
\newblock \emph{arXiv preprint arXiv:1505.05770}, 2015.

\bibitem[Rezende et~al.(2014)Rezende, Mohamed, and
  Wierstra]{rezende2014stochastic}
Danilo~Jimenez Rezende, Shakir Mohamed, and Daan Wierstra.
\newblock Stochastic backpropagation and approximate inference in deep
  generative models.
\newblock In \emph{Proceedings of the 31th International Conference on Machine
  Learning, {ICML}}, 2014.

\bibitem[Rezende et~al.(2020)Rezende, Papamakarios, Racani{\`e}re, Albergo,
  Kanwar, Shanahan, and Cranmer]{rezende2020normalizing}
Danilo~Jimenez Rezende, George Papamakarios, S{\'e}bastien Racani{\`e}re,
  Michael~S Albergo, Gurtej Kanwar, Phiala~E Shanahan, and Kyle Cranmer.
\newblock Normalizing flows on tori and spheres.
\newblock \emph{arXiv preprint arXiv:2002.02428}, 2020.

\bibitem[Sohl-Dickstein(2020)]{sohl2020two}
Jascha Sohl-Dickstein.
\newblock Two equalities expressing the determinant of a matrix in terms of
  expectations over matrix-vector products.
\newblock \emph{arXiv preprint arXiv:2005.06553}, 2020.

\bibitem[Tabak and Turner(2013)]{tabak2013family}
Esteban~G Tabak and Cristina~V Turner.
\newblock A family of nonparametric density estimation algorithms.
\newblock \emph{Communications on Pure and Applied Mathematics}, 66\penalty0
  (2):\penalty0 145--164, 2013.

\end{thebibliography}

\newpage
\appendix
\section{Experimental Details}

\label{sec:appendx}
\subsection{Cover Density}
\label{sec:app_cover_matrix}
\begin{figure}[h]
\centering
\includegraphics[width=350pt]{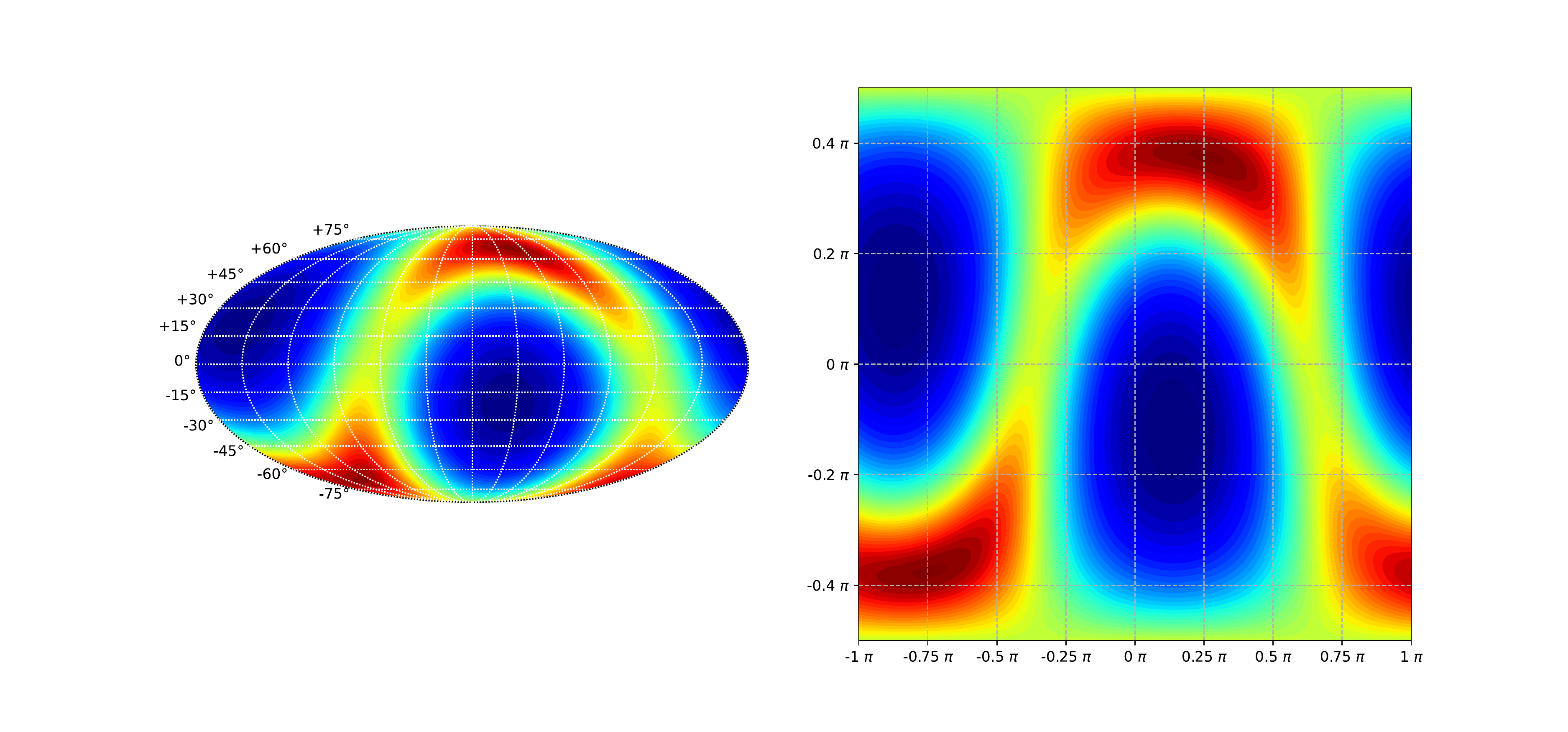}
\caption{Optimal proposal distribution corresponding to $A$.}
\label{fig:opt_proposal}
\end{figure}

The cover image of this paper illustrates the learned proposal distribution $q(s)$ corresponding to the matrix 
\begin{align*}
    A = \begin{bmatrix}
    \begin{array}{rrrr}
   -0.7056  & 0.6741 &-0.5454  \\
    0.9107  & 1.0682 & 0.1424 \\
    -1.2754  & -0.1769 & 1.0084 
    \end{array} 
\end{bmatrix},
\end{align*}
which is created with \texttt{torch.randn(3,3)} and torch manual seed $15$. The optimal proposal distribution $q^* \propto \lVert As\rVert^{-n}$ is illustrated in Figure \ref{fig:opt_proposal}. We trained the model for $10$k iterations and in contrast to the architecture in Section \ref{sec:results}, we used $N_F = 6$ flows with autoregressive masking and Neural Spline flows for both the spherical part and the interval part of $\mathbb{S}^1 \times [-1,1]$ with $N_B = 32$ bins, see again \cite{rezende2020normalizing} for details.

\subsection{Dense $10 \times 10$ Matrices}
\label{sec:app_dense_matrix}
In this section we publish the absolute determinants in Table \ref{tab:abs_dets} and the $10 \times 10$ matrices of our experiment in Section \ref{sec:dense_matrix}. Numbers are rounded to two decimals.


\begin{table}[h]
\centering
\begin{tabular}{lccccc} 
\toprule
& $A_1$ & $A_2$ & $A_3$ & $A_4$ & $A_5$  \\
\midrule 
Absolute det. & 520.36 &748.68  &945.02  &3000.5 & 252.29 \\
\bottomrule
\end{tabular}
\caption{Absolute determinants of matrices of the experiment in Section \ref{sec:dense_matrix}.}
\label{tab:abs_dets}
\end{table}

{\footnotesize
\begin{align*}
    A_1 &= \begin{bmatrix}
 \begin{array}{rrrrrrrrrr}
  -1.08 & -0.6 & 0.06 & 0.71 & -0.81 & 0.57 & 0.69 & 0.51 & -0.94 & 0.18\\
  -0.55 & 1.5 & 1.39 & -0.18 & -0.56 & -0.05 & 0.98 & 1.82 & 1.48 & 0.01\\
  -0.26 & -2.07 & -1.12 & -0.27 & -1.03 & 0.97 & -1.84 & -0.5 & -0.47 & -1.17\\
  1.01 & -1.25 & 1.71 & 1.24 & -0.79 & -0.17 & -1.05 & 0.44 & 0.02 & 0.04\\
  1.24 & -0.31 & -0.18 & -0.74 & -0.43 & 0.29 & -0.67 & 1.43 & -1.01 & -0.17\\
  -0.49 & -1.17 & 0.43 & 1.4 & 1.28 & 1.8 & -0.45 & 1.67 & -0.93 & -1.72\\
  0.78 & 1.19 & 0.02 & -0.06 & 0.72 & -1.24 & -1.19 & -0.71 & 1.73 & 0.81\\
  0.53 & 1.56 & -1.09 & 0.33 & -0.29 & -0.47 & 1.02 & 1.67 & -0.17 & 0.26\\
  1.16 & -0.18 & 0.86 & 0.94 & 0.26 & -1.64 & -0.38 & -0.31 & -0.79 & 1.31\\
  0.54 & 1.39 & -0.21 & -0.12 & 0.14 & 0.8 & 0.78 & 0.85 & -1.3 & -0.41\\
  \end{array}
\end{bmatrix} \\[3pt]
A_2 &= \begin{bmatrix}
    \begin{array}{rrrrrrrrrr}
  -1.92 & -0.19 & 0.34 & 0.41 & -0.58 & -2.08 & 0.29 & -0.46 & -1.37 & -0.45\\
  -0.56 & 0.71 & 0.06 & 0.17 & 1.44 & -1.81 & -1.19 & 1.02 & -2.84 & 2.28\\
  1.64 & 0.14 & -1.86 & 0.23 & 0.85 & 1.33 & -0.88 & -0.73 & -0.53 & 2.09\\
  -0.11 & -0.43 & 0.68 & -1.45 & 0.08 & 0.81 & 0.53 & 0.41 & 0.41 & -0.27\\
  -0.05 & 0.05 & 0.7 & -1.09 & 1.77 & -0.79 & -0.35 & 1.71 & 0.85 & 0.8\\
  1.24 & -0.22 & 0.41 & -1.02 & -0.64 & -0.21 & -1.25 & 0.71 & 0.6 & -0.75\\
  0.71 & -0.91 & -0.11 & 0.18 & 1.13 & -0.48 & 1.85 & -0.03 & 0.29 & -1.25\\
  0.52 & -1.06 & 0.48 & -2.26 & 1.52 & -0.63 & 1.26 & -1.42 & -0.02 & -1.66\\
  -1.01 & -1.23 & 0.42 & -0.37 & 1. & -0.04 & -0.32 & 0.52 & -1.91 & -1.78\\
  0.89 & -0.1 & -0.39 & -0.52 & 0.21 & -0.99 & 0.48 & 0.22 & 0.77 & -0.19\\
    \end{array}
\end{bmatrix} \\[3pt]
A_3 &= \begin{bmatrix}
  \begin{array}{rrrrrrrrrr}
  -0.15 & -1.65 & -0.95 & 0.26 & 1.35 & -0.1 & 0.37 & 0.45 & 0.23 & -1.12\\
  0.61 & -1.81 & -0.68 & 0.58 & 0.94 & 2.36 & -0.49 & 0.04 & 0.86 & 0.52\\
  1.91 & -1.44 & -0.51 & 0.96 & -2.56 & -0.01 & -1.13 & 0.19 & -2.5 & 0.68\\
  0.93 & -1.3 & -0.65 & -1.9 & -0.09 & 0.24 & 0.69 & 1.28 & -0.57 & -0.39\\
  0.55 & -1.34 & -1.37 & -1.29 & -0.28 & -0.67 & 0.77 & -0.25 & 0.85 & -2.89\\
  -0.86 & 1.95 & -1.33 & 0.68 & 0.27 & 0.25 & 0.29 & -1.29 & 2.05 & 0.11\\
  0.82 & 0.52 & -0.71 & -0.59 & -1.57 & -1.05 & 0.46 & -0.61 & 0.63 & 2.02\\
  0.76 & 0.01 & -0.06 & -0.43 & 1.12 & 1.05 & -1.35 & -0.04 & -0.62 & -0.35\\
  -2.13 & -0.8 & 1.12 & 1.77 & -0.79 & -0.1 & 1.17 & -1.06 & -0.37 & 0.01\\
  -0.86 & 1.44 & -0.55 & 1.19 & 2.52 & 0.81 & -0.36 & -0.61 & 1.24 & -0.06\\
    \end{array}
\end{bmatrix} \\[3pt]
A_4 &= \begin{bmatrix}
  \begin{array}{rrrrrrrrrr}
  -0.31 & -0.68 & -0.22 & -0.28 & 1.64 & -0.41 & -0.66 & -0.59 & 1.57 & 0.38\\
  -0.15 & 0.6 & 1.08 & 1.29 & -0.12 & 1.89 & -1.85 & -0.11 & 1.5 & 0.72\\
  0.88 & -1.71 & 0.69 & -1.75 & -0.06 & 0.9 & 0.08 & -0.11 & -0.21 & 1.75\\
  0.31 & -1.4 & -1.79 & 0.17 & 0.57 & -0.86 & 1.64 & -1.55 & 0.91 & -2.06\\
  1.1 & -1.19 & 0.47 & -0.84 & 0.37 & 0.25 & 0.03 & -0.23 & 1.32 & 0.36\\
  0.43 & -0.02 & -0.04 & 1.19 & 0.2 & -1.13 & 1.36 & 1.23 & -0.01 & 2.08\\
  -0.8 & 0.48 & -1.57 & 0.6 & -0.19 & -0.18 & -0.88 & -1.53 & -0.66 & -0.83\\
  1.32 & -1.09 & 0.71 & 1.04 & 1.02 & -0.09 & 1.51 & -0.51 & -0.73 & -0.82\\
  0.21 & -2.07 & 0.61 & 0.29 & 1.41 & -1.93 & -2.06 & 0.23 & -0.09 & 0.24\\
  -1.36 & 0.3 & 0.15 & 1.33 & -1.1 & -0.72 & 0.37 & 0.09 & -0.56 & 2.81\\
    \end{array}
\end{bmatrix} \\[3pt]
A_5 &= \begin{bmatrix}
  \begin{array}{rrrrrrrrrr}
  0.36 & -0.95 & 0.12 & 0.85 & -0.4 & -0.1 & 0.58 & -0.48 & 0.79 & 0.12\\
  0.11 & -0.02 & -0.66 & -0.98 & -0.28 & -1.61 & -0.82 & 1.13 & 1.18 & 0.33\\
  -0.7 & 0.65 & -1.5 & -0.33 & -0.18 & -0.6 & -0.84 & -0.43 & -0.42 & 1.12\\
  -2.33 & -0.49 & 0.61 & 0.88 & -0.85 & -0.68 & 0.38 & 0.53 & 0.34 & 1.59\\
  0.43 & 1.61 & -0.14 & 1.15 & -1.25 & 2.28 & -0.32 & -0.36 & -2.1 & 0.98\\
  -0.68 & -0.54 & -0.88 & 1.55 & 0.7 & -1.34 & 0.15 & -0.27 & -0.86 & 1.35\\
  -0.83 & -0.52 & -0.83 & -1.98 & 1.79 & -0.86 & 0.05 & 1.29 & 0.1 & 1.17\\
  -1.34 & -0.66 & 0.12 & -0.95 & -0.46 & 2.15 & -0.67 & -0.77 & 1.87 & 1.4\\
  0.54 & -0.51 & 0.16 & 1.38 & 1.49 & 0.61 & 0.22 & 0.64 & -0.27 & -0.47\\
  0.62 & -0.24 & -0.11 & 0.27 & -0.48 & 0.75 & 0.59 & 0.41 & -0.81 & 0.07\\
    \end{array}
\end{bmatrix}
\end{align*}
}

\subsection{Filter of Convolutional Layer}
\label{sec:app_conv_layer}
Here we present the $3 \times 3$ filter $k$ which was used to create the $16 \times 16$ matrix $W$ of our experiment in section \ref{sec:conv_layer}. The matrix $W$ is the equivalent matrix of the convolution $k \star x$ where $x$ is an arbitrary $4 \times 4$ image.
\small{
\begin{align*}
    k = \begin{bmatrix}
    \begin{array}{rrr}
  -0.107 & -0.689 & -0.027\\
  0.226 & 1.393 & -0.544\\
  -0.28 & -0.467 & 0.024\\
  \end{array}
\end{bmatrix}
\end{align*}
}

\end{document}